\documentclass[twoside,11pt]{article}

\usepackage{blindtext}

\usepackage{jmlr2e}

\usepackage{lastpage}

\usepackage{setspace}
\usepackage{graphicx,subcaption}
\usepackage{xcolor}

\usepackage{amsmath}
\usepackage{amssymb}
\usepackage{bbm}
\usepackage{systeme}

\ShortHeadings{Random Projection Forest Initialization for Graph Convolutional Networks}{Random Projection Forest Initialization for Graph Convolutional Networks}
\firstpageno{1}

\begin{document}

\title{Random Projection Forest Initialization for Graph Convolutional Networks}

\author{\name Mashaan Alshammari \email mashaan.awad1930@alum.kfupm.edu.sa \\
       \addr Independent Researcher\\
       Riyadh, Saudi Arabia
       \AND
       \name John Stavrakakis \email john.stavrakakis@sydney.edu.au \\
       \addr School of Computer Science\\
       The University of Sydney\\
       NSW 2006, Australia
       \AND
       \name Adel F. Ahmed \email adelahmed@kfupm.edu.sa \\
       \addr Information and Computer Science Department\\
       King Fahd University of Petroleum and Minerals\\
       Dhahran, Saudi Arabia
       \AND
       \name Masahiro Takatsuka \email masa.takatsuka@sydney.edu.au \\
       \addr School of Computer Science\\
       The University of Sydney\\
       NSW 2006, Australia}

\editor{}

\maketitle

\begin{abstract}
  Graph convolutional networks (GCNs) were a great step towards extending deep learning to graphs. GCN uses the graph $G$ and the feature matrix $X$ as inputs. However, in most cases the graph $G$ is missing and we are only provided with the feature matrix $X$. To solve this problem, classical graphs such as $k$-nearest neighbor ($k$-nn) are usually used to construct the graph $G$ and initialize the GCN. Although it is computationally efficient to construct $k$-nn graphs, the constructed graph might not be very useful for learning. In a $k$-nn graph, points are restricted to have a fixed number of edges, and all edges in the graph have equal weights. Our contribution is Initializing GCN using a graph with varying weights on edges, which provides better performance compared to $k$-nn initialization. Our proposed method is based on random projection forest (rpForest). rpForest enables us to assign varying weights on edges indicating varying importance, which enhanced the learning. The number of trees is a hyperparameter in rpForest. We performed spectral analysis to help us setting this parameter in the right range. In the experiments, initializing the GCN using rpForest provides better results compared to $k$-nn initialization.
\end{abstract}


\section{Introduction}
\label{Introduction}

Convolutional neural networks (CNNs) proved to be effective in many applications. The convolution component in CNNs is only applicable to fixed grids like images and videos. Applying CNNs to non-grid data (graphs for example) can be useful for many applications. Graph Convolutional Networks (GCNs) \cite{kipf2017semi} introduced a convolution component designed for graphs, where vertices are allowed to have a varying number of neighbors unlike fixed grids. GCNs were used in several application such as sentiment analysis \cite{Phan2023Aspect}, computer vision tasks \cite{Ren2022Graph} and ranking gas adsorption properties \cite{cong2022prediction}.

Because the edges arrangements are different from one graph to another, GCN needs the adjacency matrix $A$ to perform the convolutions. The GCN performs two changes to the input adjacency matrix: 1) it adds self-loops to include the vertex own feature vector into the convolution, and 2) it normalizes the adjacency matrix using the degree matrix $D$ to avoid favoring vertices with many edges. These two changes were embedded in the convolution function of GCN. Researchers have improved the GCN algorithm in different ways such as: increasing the depth of GCN \cite{Li2023DeepGCNs}, implementing attention mechanism \cite{velickovic2018graph,cong2022prediction}, or replacing self-loops with trainable skip connection \cite{Tsitsulin2023Graph}. Our focus is on creating an adjacency matrix in an unsupervised way.

GCN can classify graph vertices efficiently if the adjacency matrix is given, but it cannot create an adjacency matrix from scratch. This opens a new research track on how to create an adjacency matrix for GCN. An obvious solution is to use traditional methods to create the adjacency matrix such as: fully connected graph, $k$-nearest neighbor graph, and $\epsilon$-neighborhood graph \cite{Luxburg2007tutorial}. The fully connected graph grows exponentially with the number of vertices ($n$), which makes $k$-nn and $\epsilon$-graphs more appealing. But if we compare $k$-nn and $\epsilon$-graphs, we can see that $k$-nn graphs can be implemented using efficient data structures such as $kd$-trees.

Franceschi et al. proposed a bi-level optimization for GCN graphs \cite{franceschi2019learning}. They used a number of sample graphs to train the GCN and based on the validation error they modified the original adjacency matrix. Although the method proposed by Franceschi et al. is effective in learning the adjacency matrix, it still uses the $k$-nearest neighbor graph. $k$-nn graphs have two problems. First, all vertices in a $k$-nn graph are restricted to $k$ edges, this would limit the ability of vertices to connect to more similar neighbors. Second, all edges in a $k$-nn graph are assigned equal weights, which gives them the same level of importance when passed through a GCN.

We present a new method to construct the adjacency matrix for GCN. We used random projection forests (rpForest) \cite{Yan2018Nearest,Yan2021Nearest} to construct the adjacency matrix. An rpForest is a collection of random projection trees (rpTree). rpTrees use random directions to partition the data points into tree nodes \cite{Dasgupta2008Random,Dasgupta2015Randomized,Keivani2021Random}. A leaf node in an rpTree represents a small region that contains similar points. We connect all points in a leaf node in each rpTree. If an edge keeps persisting over multiple rpTrees it will be assigned a higher weight. This would solve the equal weights problem in $k$-nn graphs. Also, since leaf nodes can have a varying number of points, this would allow points to connect to more or less neighbors depending on the density inside the leaf node. The experiments showed that a GCN with rpForest initialization performed better than a GCN with $k$-nn initialization. Our contributions are:

\begin{itemize}
	\item Initializing GCN using a graph based on rpForest, that allows points to connect to a varying number of neighbors and assigns weights proportionate to the edge's occurrence in the rpForest.
	\item Providing a spectral analysis to set the hyperparameter (number of trees $T$) in rpForest.
\end{itemize}

\section{Related work}
\label{RelatedWork}

This section discusses the recent advancements in graph convolutional networks, graph construction methods, and random projection forests. These three topics form the basis of our proposed method.

\subsection{Graph Convolutional Networks (GCNs)}

The successful application of convolutional neural network (CNN) on imagery data has stimulated research to extend the convolution concept beyond images. Images can be viewed as a special case of graphs where edges and vertices are ordered on a fixed grid. The problem with applying convolutions on graphs is that vertices have a varying number of edges. GCN extended the convolution to graphs by performing three steps: 1) feature vectors are averaged within the node’s local neighborhood, 2) the averaged features are transformed linearly, and 3) a nonlinear activation is applied to the averaged features \cite{Wei2023Adaptive}.

Wu, et al. \cite{Wu2021Comprehensive} have categorized graph convolutional network methods into two categories: 1) spectral-based GCNs and 2) spatial-based GCNs. Spectral-based GCNs rely on spectral graph theory. The intuition is that the graph Laplacian carries rich information about graph geometry. A symmetric graph Laplacian is defined as $L=D^{-\frac{1}{2}}LD^{-\frac{1}{2}}=I-D^{-\frac{1}{2}}AD^{-\frac{1}{2}}$ \cite{Ng2001Spectral,Luxburg2007tutorial}. $D$ is the degree matrix in which the diagonal shows the degree of each vertex $D_{ii}=\sum_{j} A_{ji}$. The graph Fourier transform projects an input graph signal $x$ to an embedding space, where the basis are formed by eigenvectors of the normalized graph Laplacian $L$. The process of graph convolution can be thought of as convoluting an input signal $x$ with a filter $g_\theta = diag(\theta)$ as shown in Equation (\ref{Eq-001}): 

\begin{equation}
	g_\theta * x = U g_\theta U^\top x ,
	\label{Eq-001}
\end{equation}
\noindent
where $\theta \in \mathbb{R}^{n}$ is a parameter that controls the filter $g$ and $U$ is the matrix of eigenvectors ordered by eigenvalues. Bruna, et al. represented the filter $g$ as a set of learnable parameters \cite{Bruna2013Spectral}. Henaff, et al. \cite{Henaff2015Deep} extended the model proposed by Bruna, et al. \cite{Bruna2013Spectral} to datasets where graphs are unavailable. Wu, et al. \cite{Wu2021Comprehensive} have identified three limitations for spectral-based GCNs: 1) any change in the graph structure would change the embedding space, 2) the learned filters are domain specific and cannot be applied to different graphs, 3) they require an eigen decomposition step, which is computationally expensive $O(n^3)$.

Although the graph spectrum provides rich information, avoiding the eigen decomposition step will be a huge boost in terms of performance. Hammond, et al. \cite{Hammond2011Wavelets} proposed an approximation of the filter $g_\theta$ via Chebyshev polynomials $T_k(x)$, which was deployed into GCN by Defferrard, et al. \cite{Defferrard2016Convolutional}. The Chebyshev polynomials are recursively defined as:

\begin{equation}
	T_k(x) = 2x T_{k-1}(x) - T_{k-2}(x),
	\label{Eq-002}
\end{equation}
\noindent
with $T_0(x)=1$ and $T_1(x)=x$. This formulation would allow us to perform the convolution on graphs as:
\begin{equation}
	g_{\theta'} * x = \sum_{k=0}^{K} \theta'_k T_k(\tilde{L})x ,
	\label{Eq-003}
\end{equation}
\noindent
with $\tilde{L} = \frac{2}{\lambda_{max}} L - I_N$. This $K$ localization is a $K^{th}$-order neighborhood, which means that it depends on the nodes that are $K$ steps away from the central node. This graph convolution was simplified by Kipf and Welling \cite{kipf2017semi}. They limited the layer-wise convolution operation to $K = 1$, and set $\lambda_{max} = 2$. Using these settings, the graph convolution in Equation (\ref{Eq-003}) can be simplified to:

\begin{equation}
	g_{\theta'} * x = \theta_0 x - \theta_1 D^{-\frac{1}{2}}AD^{-\frac{1}{2}} x .
	\label{Eq-004}
\end{equation}
\noindent
The time complexity for GCN is $O(LAd + LN{d^2})$, where $L$ is the number of layers, $A$ is the adjacency matrix, $d$ is the number of features, and $N$ is the number of nodes \cite{You2020GCN}.

\subsection{Graph Construction}

All methods explained in the previous section assume the graph $G(V,E)$ to be already constructed. But this is not the case in many practical applications, where only the feature matrix $X$ is provided. When a graph is missing, the most used way to construct it is to use the Gaussian heat kernel \cite{Belkin2001Laplacian}:

\begin{equation}
	A_{ij}=\exp{\left(\frac{-d^2\left(i,j\right)}{2\sigma^2}\right)} ,
	\label{Eq-005}
\end{equation}
\noindent
where $d^2\left(i,j\right)$ is the distance between the samples $i$ and $j$. The problem with the heat kernel is that it heavily depends on the selection of the scaling parameter $\sigma^2$, and usually the user has to try different values and selects the best one. This was improved by the self-tuning diffusion kernel \cite{Zelnik2004Self}:

\begin{equation}
	A_{ij}=\exp{\left(\frac{-d^2\left(i,j\right)}{\sigma_i\ \sigma_j}\right)},
	\label{Eq-006}
\end{equation}
\noindent
where $\sigma_i$ is the distance from the point $x_i$ to its $K^{th}$ neighbor. Zelnik-manor and Perona have set $K=7$ \cite{Zelnik2004Self}.

Constructing a graph using these two approaches requires performing pairwise comparisons, which means computations in order of $O(n^2)$. To avoid these computations, one could use a more efficient data structure. $k$-nearest neighbor graphs are usually implemented using k-dimensional trees (also known as $k$d-trees) \cite{Bentley1975Multidimensional}. $k$d-trees start by selecting the dimension with maximum dispersion. Along that dimension they split at the median and place whatever less than the median in the left child and whatever greater than the median in the right child. After several recursive executions, the $k$d-tree algorithm scans the leaf nodes and returns the $k$-nearest neighbors. The $k$-nearest neighbor graph is defined as:

\begin{equation}
	A_{ij} = 
	\systeme*{
		&1 \text{,} \quad j\in knn(i),
		&0 \text{,} \quad \text{otherwise}		
	} .
	\label{Eq-007}
\end{equation}
\noindent
We can identify two problems with a $k$-nearest neighbor graph: 1) it assigns equal weights on edges which gives all edges the same importance, and 2) it restricts all points to have $k$ edges regardless of their position in the feature space.

\begin{figure}[]
	\centering
	\includegraphics[width=0.75\textwidth,height=20cm,keepaspectratio]{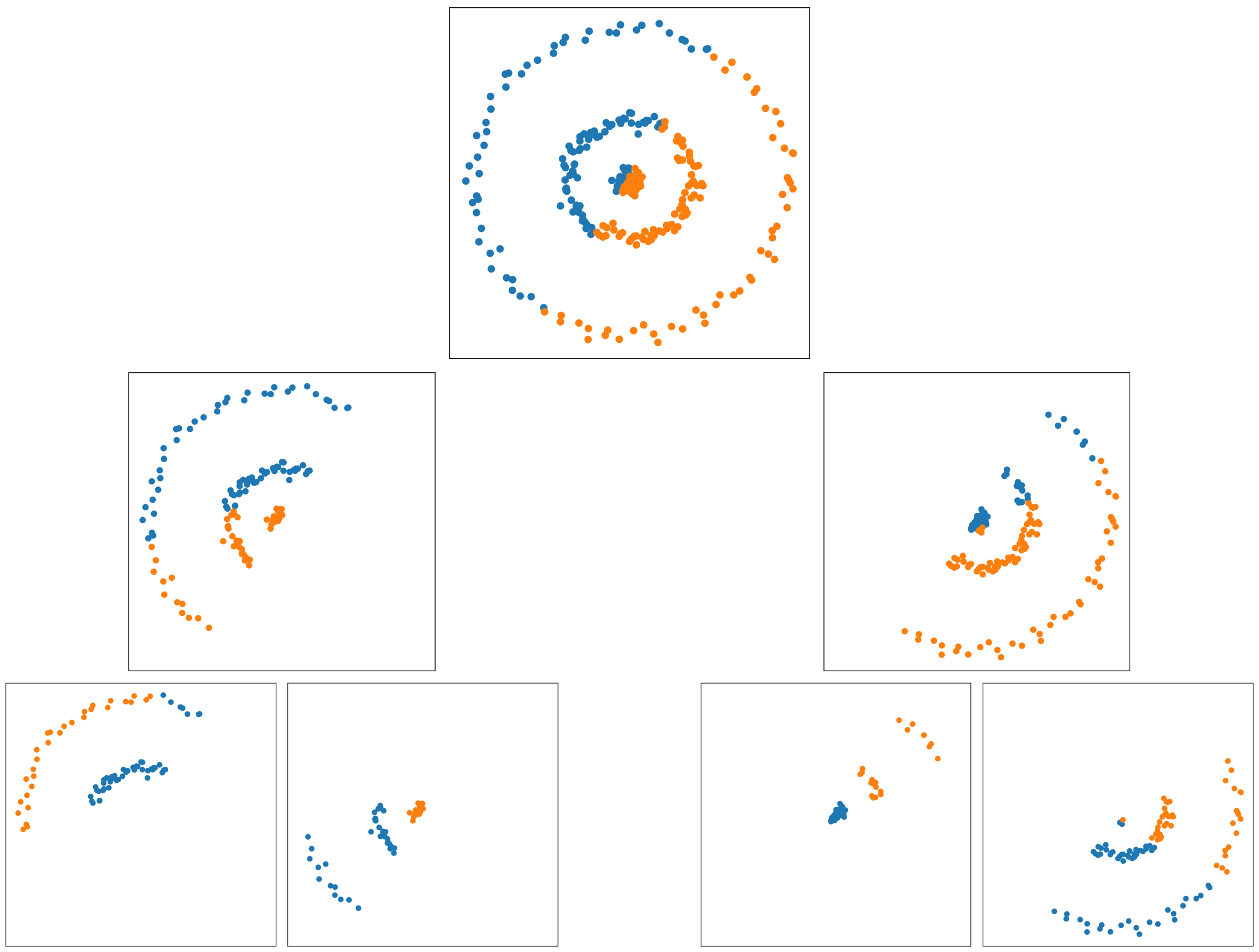}	
	\caption{An example of running rpTree algorithm on points in 2D. At each node in the tree, a random direction is selected, and all points are projected onto it. Points are split at the median, where points less than the median (points in blue) are placed in the left child, and points larger than the median (points in orange) are placed in the right child. (Best viewed in color)}
	\label{Fig:rpTree}
\end{figure}

\subsection{Random Projection Forests (rpForests)}

Random projection forest (rpForest) is a collection of random projection trees (rpTrees) \cite{Dasgupta2008Random,Freund2008Learning}. rpTrees use the same principle as $k$d-trees, that is partitioning the feature space and placing points in a binary tree. The difference is that $k$d-tree splits along the existing dimensions, while rpTree splits along random directions. In rpTrees, the root node contains all the data points, and the leaf nodes contain disjoint subsets of these data points. Each internal node in rpTree holds a random projection direction $\overrightarrow{r}$ and a scalar split point $c$ along that random direction. Figure\ \ref{Fig:rpTree} shows an example of rpTree.

The most common use of rpTrees is performing $k$-nearest neighbor search. Yan, et al. \cite{Yan2021Nearest} used rpForest (i.e., a collection of rpTrees) to perform $k$-nn search. They modified the splitting rule by selecting three random directions, then project onto the one that yields the maximum dispersion of points. rpTree was used in anomaly detection application by Chen, et al. \cite{Chen2015Anomaly}. They also modified the splitting rule by examining if the points form two Gaussian components, they will split these two components into left and right nodes. Tavallali, et al. \cite{Tavallali2021Kmeans} proposed a $k$-means tree, which outputs the centroids of clusters. All these modifications on rpTree are supported by the empirical evidence provided by Ram, et al. \cite{Ram2013Which}. They stated that the best performing binary space-partitioning trees are the ones that have better vector quantization and large partition margins. But these modifications add extra computations to the rpTree algorithm.

\section{Proposed Method}
\label{ProposedApproach}

Our proposed method has two components: a neural network and a graph construction component. The next section explains the neural network component, followed by two sections discussing the graph construction component.

\subsection{GCN and LDS}
Graph convolutional networks (GCNs) are used for semi-supervised node classification. A GCN propagation rule at the first layer is defined as:

\begin{equation}
	f(X,A) = ReLU\left(AXW^{(0)}\right),
	\label{Eq-008}
\end{equation}
\noindent
where $A$ is the adjacency matrix, $ReLU$ is the activation function $f(x)=max(0,x)$, and $W^{(0)}$ is the weight matrix for the first neural network layer. Two problems arise from this definition. First, the node's own feature vector is not included since $A$ has zeros on the diagonal. This can be solved by allowing self-loops, and rewrite the adjacency matrix to be $\tilde{A}=A+I$. The second problem is the normalization of the adjacency matrix, which can be solved by normalizing using the degree matrix $D$. The adjacency matrix becomes $\hat{A}=D^{-\frac{1}{2}}\tilde{A}D^{-\frac{1}{2}}$. By applying these changes, the GCN propagation rule in Equation (\ref{Eq-008}) can be rewritten as:

\begin{equation}
	f(X,A) = ReLU\left(\hat{A}XW^{(0)}\right).
	\label{Eq-009}
\end{equation}
\noindent
The most common architecture is a two-layer GCN, which can be defined using the following formula:

\begin{equation}
	f(X,A) = softmax\left(\hat{A}\ ReLU\left(\hat{A}XW^{(0)}\right)W^{(1)}\right).
	\label{Eq-010}
\end{equation}

Franceschi, et al. \cite{franceschi2019learning} proposed LDS method to learn the adjacency matrix $A$. It stands for Learning Discrete Structures (LDS). They used $k$-nearest neighbor graph to initialize the Graph Convolutional Networks (GCNs). The methodology involves four steps: initialization, sampling, inner optimization, and outer optimization. First, a parameter $\theta$ is initialized to be the adjacency matrix of $k$-nn graph and run GCN once to initialize its parameters. Then, the method iteratively sample graphs from $\theta$ to optimize for GCN parameters (inner optimizer) and $\theta$ (outer optimizer) in a bilevel optimization.

In our experiments, we used both methods GCN and LDS to evaluate their performance when we initialize them using different graphs. The initializations we used in the experiments are $k$-nearest neighbor graph initialization and random projection forest (rpForest) initialization.

\subsection{$k$-nearest neighbor Initialization}

GCN needs a graph to perform the convolutions. A common choice is to use the $k$-nearest neighbor graphs to initialize the GCN. In a $k$-nn graph, each point is connected to its nearest $k$ neighbors. Intuitively, the adjacency matrix contains $k \times n$ nonzero entries since we have $n$ points and each one of them has $k$ edges. A formal definition for $k$-nn graphs is given in Equation (\ref{Eq-007}).

\begin{figure}
	\centering
	\includegraphics[width=0.40\textwidth,height=20cm,keepaspectratio]{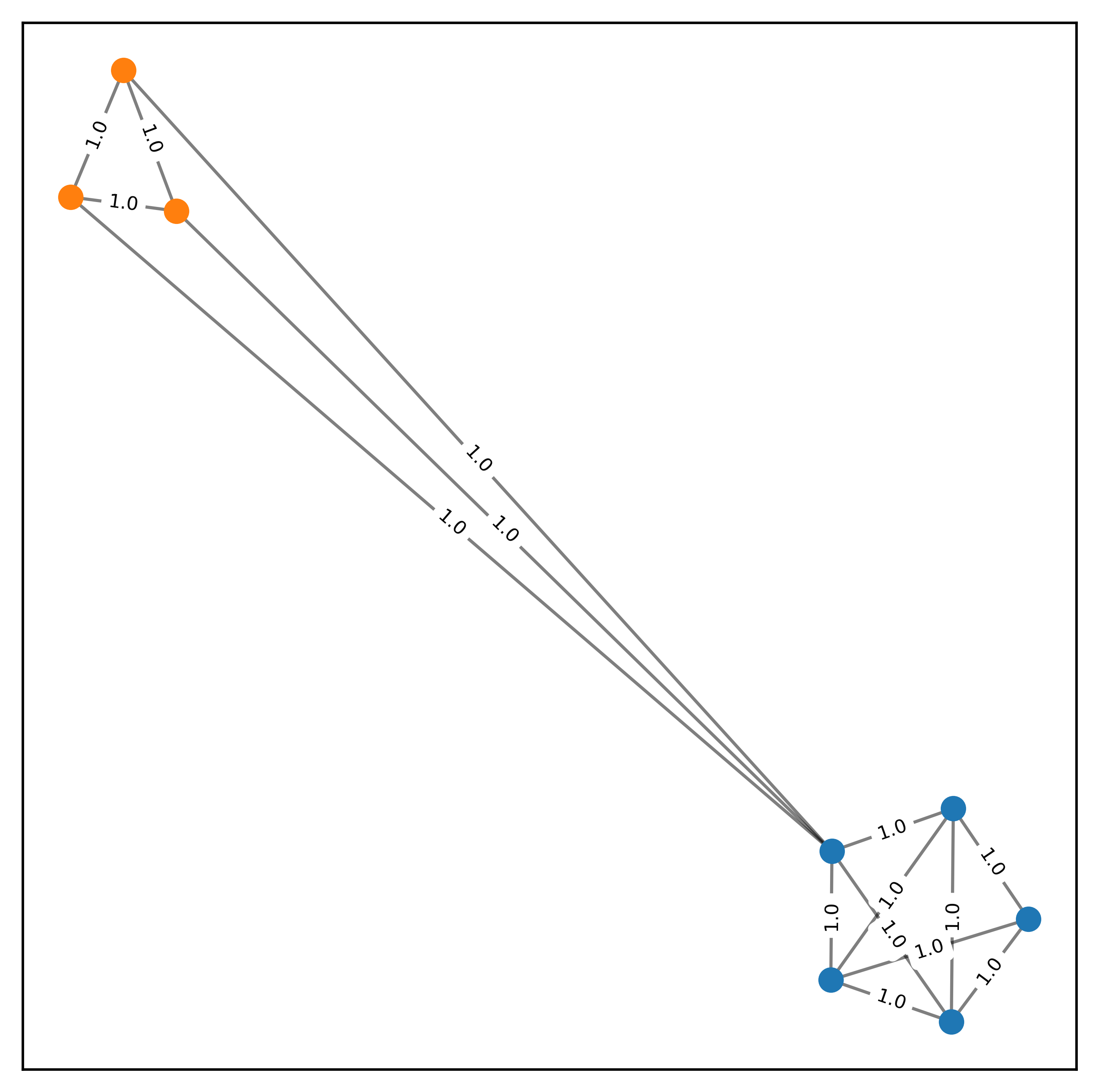}	
	\caption{A $k$-nn graph with $k=3$; all edges were assigned equal weights even the ones connecting two different classes. (Best viewed in color)}
	\label{Fig:graph-knn}
\end{figure}

An example of $k$-nearest neighbor graph is shown in Figure\ \ref{Fig:graph-knn}. In that figure, we have three points in the blue class and five points in the orange class. The graph was constructed with $k=3$. Note that some edges connect two points from two different classes. The existence of these edges is undesirable because they could confuse the classifier. But in an unsupervised graph construction, these edges are sometimes unavoidable. The solution is to assign a small weight on these edges connecting two different classes. Unfortunately, we cannot do that in $k$-nn graphs because all edges get an equal weight.

$k$-nn provides a fast initialization for the graph convolution networks (GCN). But it assigns equal weights to all edges which gives the edges spanning two classes the same importance as edges connecting one class. To avoid this problem, we need a graph construction scheme that assigns small weights on edges connecting two classes.

\subsection{rpForest Initialization}

The random projection tree (rpTree) is a binary space-partitioning tree. The root node contains all points in the dataset. For each node in the tree, the method picks a random direction $\overrightarrow{r}$. The dimensions of $\overrightarrow{r}$ is $\mathbb{R}^{d-1}$, where $d$ is the number of dimensions in the dataset. All points in the tree node get projected onto $\overrightarrow{r}$. Then, a split point $c$ is selected randomly between $\lbrack \frac{1}{4}, \frac{3}{4} \rbrack$ along $\overrightarrow{r}$. In the projection space, if a point is less than $c$ it is placed in the left child, otherwise it is placed in the right child. rpTrees are particularly useful for $k$-nearest neighbor search. But in this paper, we are going to use them for graph construction.

\begin{figure}
	\centering
	\includegraphics[width=0.99\textwidth,height=20cm,keepaspectratio]{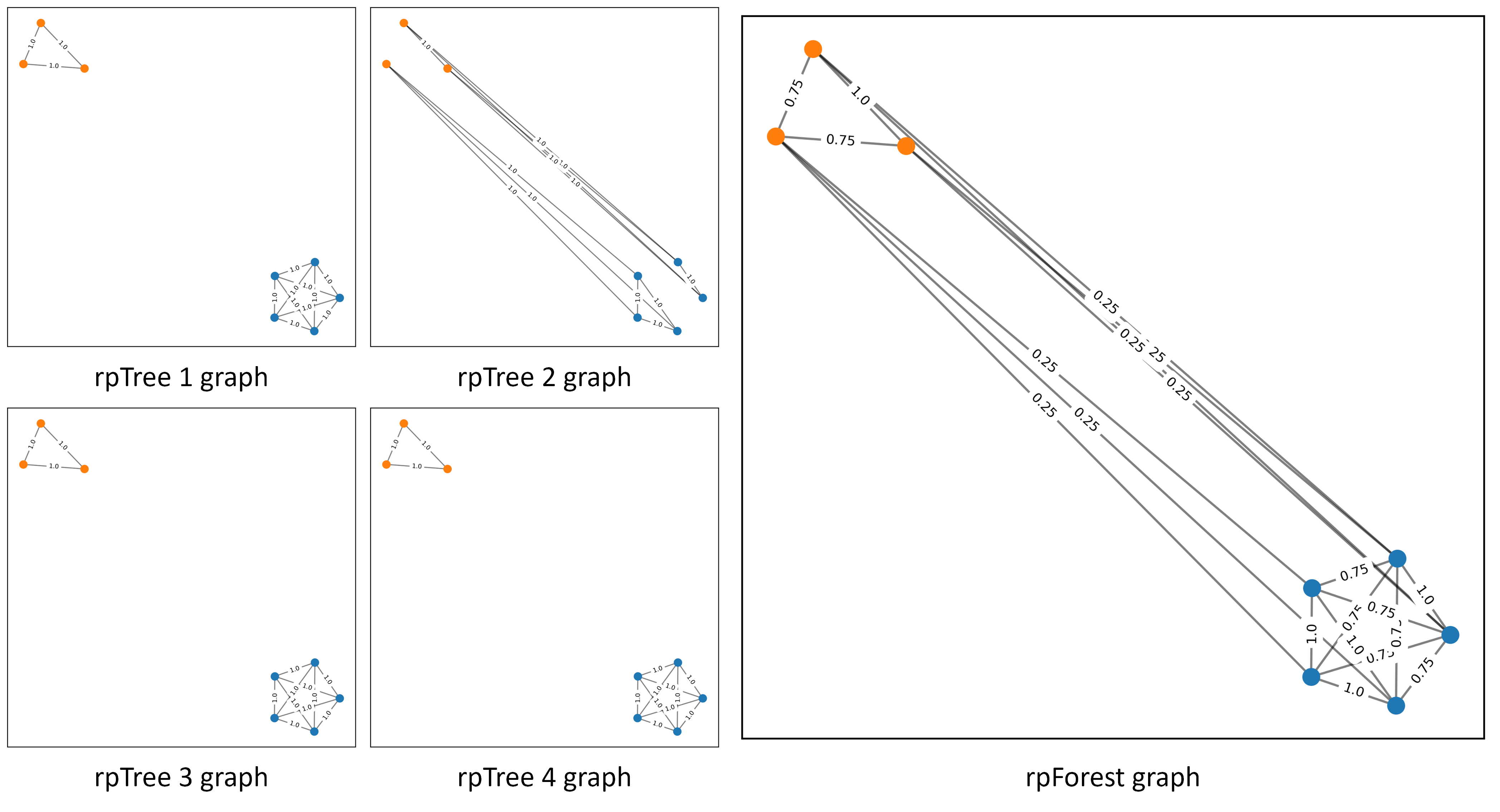}	
	\caption{An rpForest graph with $T=4$; edges connecting the two classes were assigned a small weight (0.25) because they only appear in one tree out of four. (Best viewed in color)}
	\label{Fig:graph-rpForest}
\end{figure}

A collection of rpTrees is called rpForest. rpForests were proposed by Yan, et al. \cite{Yan2018Nearest,Yan2021Nearest} and they applied it in spectral clustering similarity and $k$-nearest neighbor search. We used rpForests to construct a graph and use it as an input to the GCN. rpForests helped us to overcome two problems we identified with $k$-nn graphs. The problem of equal weights on edges, and the problem of restricting points to a fixed number of neighbors. We constructed a number of rpTrees. Then, we connect all points falling into the same leaf node. The intuition is simple, if a pair of points fall into the same leaf node in all rpTrees, they will be connected with the maximum weight. Otherwise, the weight on the edge connecting them will be proportionate to the number of the leaf nodes they fall in together. Also, the points will not be restricted to a fixed number of edges because the number of points varies from one leaf node to another. 

Figure\ \ref{Fig:graph-rpForest} illustrates how did we construct a graph using rpForest. In that figure we used four rpTrees each of which has two levels meaning we only perform the split once. Apart from (rpTree 2), all trees have split the two classes into two different leaf nodes. The final graph aggregates all edges in the trees. The edges connecting the two classes were assigned a small weight (0.25) because they only appear in one tree out of four. Edges connecting points from the same class were assigned higher weights (1 or 0.75) because they either appear in all trees or in three out of four trees.

\section{Experiments and discussions}
\label{Experiments}

We designed our experiments to test how $k$-nn and rpForest graphs affect the performance of GCN \cite{kipf2017semi} and LDS \cite{franceschi2019learning}. Unlike GCN, LDS iteratively improves the original graph based on the validation error. For the $k$-nn graph, we used the same settings used by the LDS algorithm, where $k$ was set to be $10$. For the rpForest graph, we set the number of trees $T$ to be $10$. In the next section we provide an empirical examination showing why it is safe to use $T=10$ as the number of trees. The number of layers in GCN was kept less than 4 layers, to prevent a drop in the performance \cite{Li2023DeepGCNs}. We used two evaluation metrics: 1) the test accuracy was used to evaluate the performance, and 2) the number of edges in the graph was used to evaluate the storage efficiency.

We modified the original python files provided by GCN and LDS, to include the code for rpForest graph. We used a different font for the names of datasets. The name of a dataset is written as \texttt{dataset}. The code used to produce the experiments is available on \url{https://github.com/mashaan14/RPTree-GCN}. All experiments were coded in python 3 and run on a machine with 20 GB of memory and a 3.10 GHz Intel Core i5-10500 CPU.

\begin{figure}[h]
	\centering
	\includegraphics[width=0.70\textwidth,height=20cm,keepaspectratio]{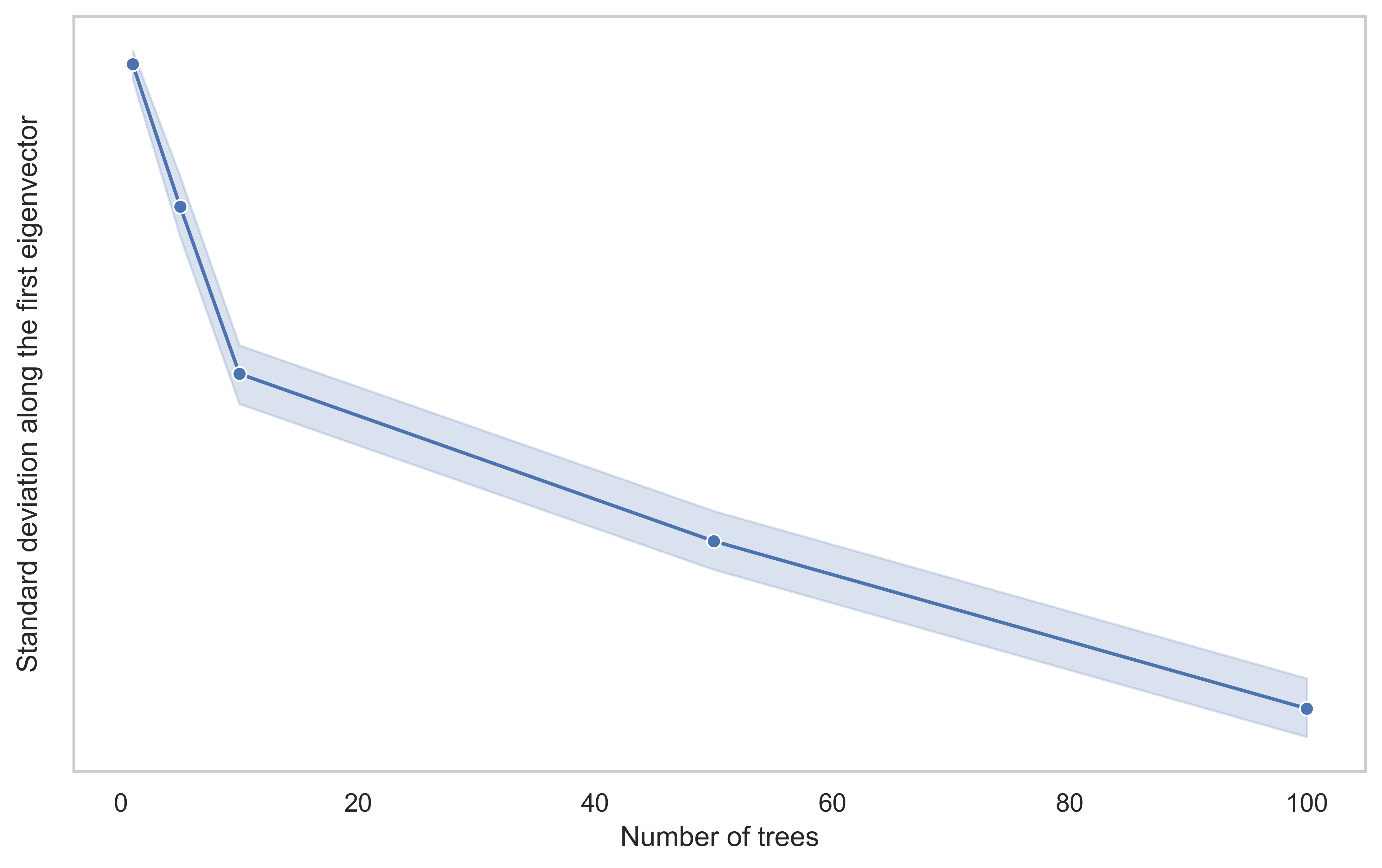}	
	\caption{Measuring the standard deviation of points along the smallest eigenvector $v_0$; $T=10$ represents an elbow point. (Best viewed in color)}
	\label{Fig:EigenvectorStd}
\end{figure}

\subsection{Using spectral analysis to set the number of trees}

We are constructing the graph out of the leaf nodes in the rpForest. The number of rpTrees $T$ is a hyperparameter in rpForest. Tuning $T$ highly influences the outcome of rpForest. We can identify two problems that could occur from different values of $T$. The first problem occurs when $T$ is set to a low value, which risks feeding a disconnected graph to the GCN. A disconnected graph could mean one of the classes is not connected, which negatively affects the performance of the GCN. The second problem occurs when $T$ is set to a high value. This will lead to a graph with so many edges, which could affect the memory efficiency of our method.

\begin{figure}[]
	\begin{subfigure}{0.24\textwidth}
		\centering
		\includegraphics[width=\linewidth]{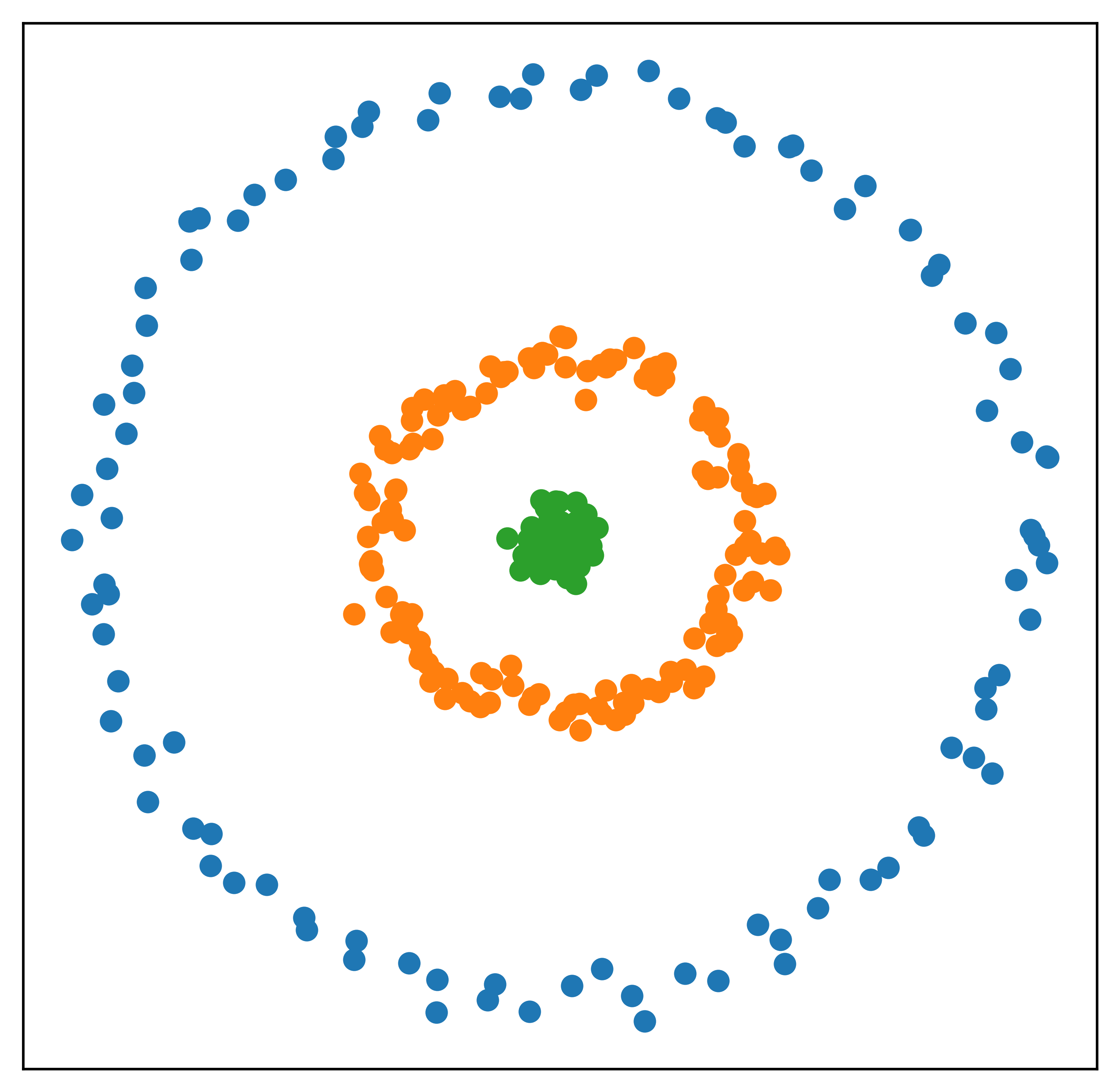}
		\caption{\texttt{3rings299}}
		\label{Fig:Fig-04-a}
	\end{subfigure}  
	\begin{subfigure}{0.24\textwidth}
		\centering
		\includegraphics[width=\linewidth]{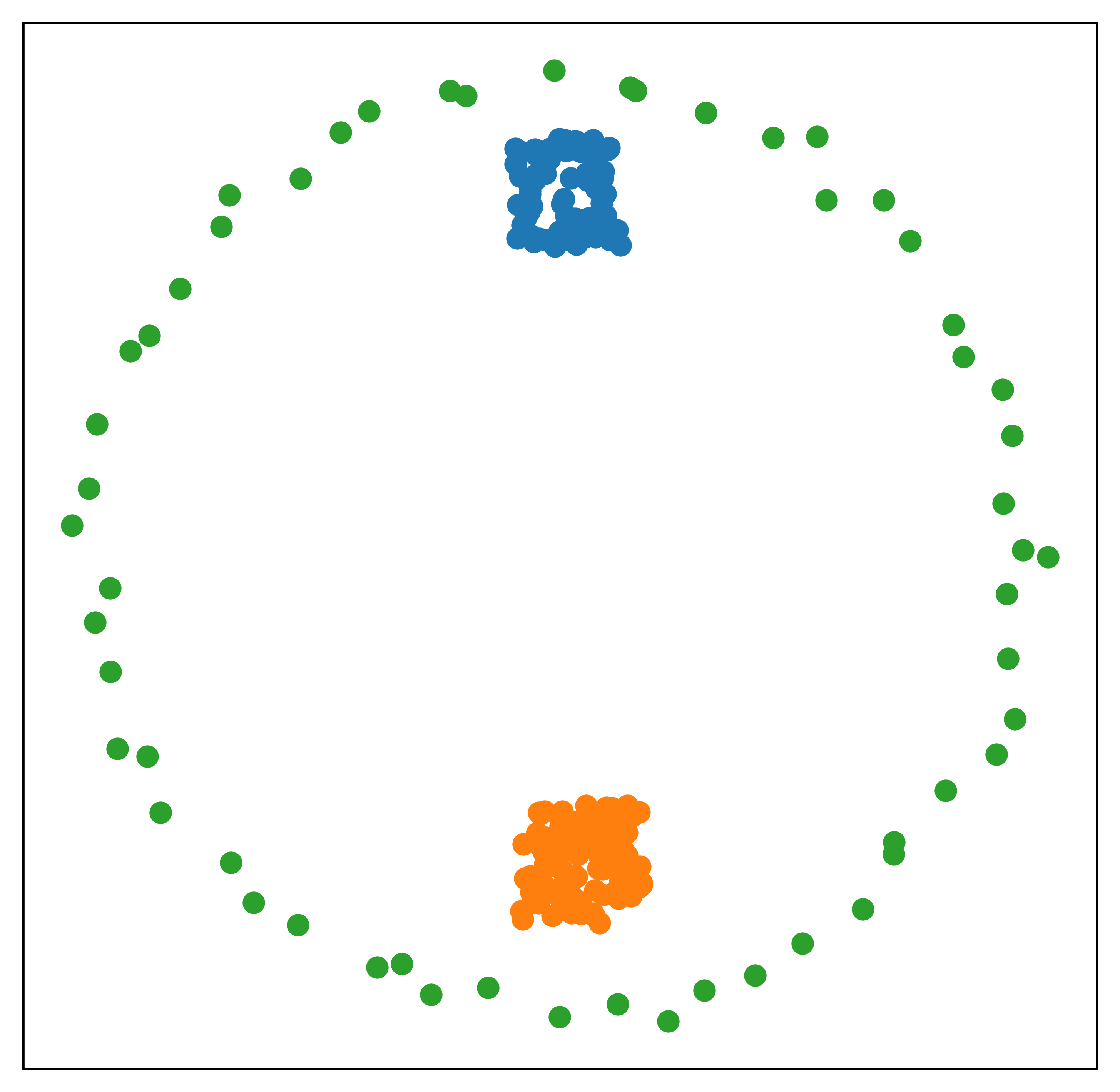}
		\caption{\texttt{ring238}}
		\label{Fig:Fig-04-b}
	\end{subfigure}
	\begin{subfigure}{0.24\textwidth}
		\centering
		\includegraphics[width=\linewidth]{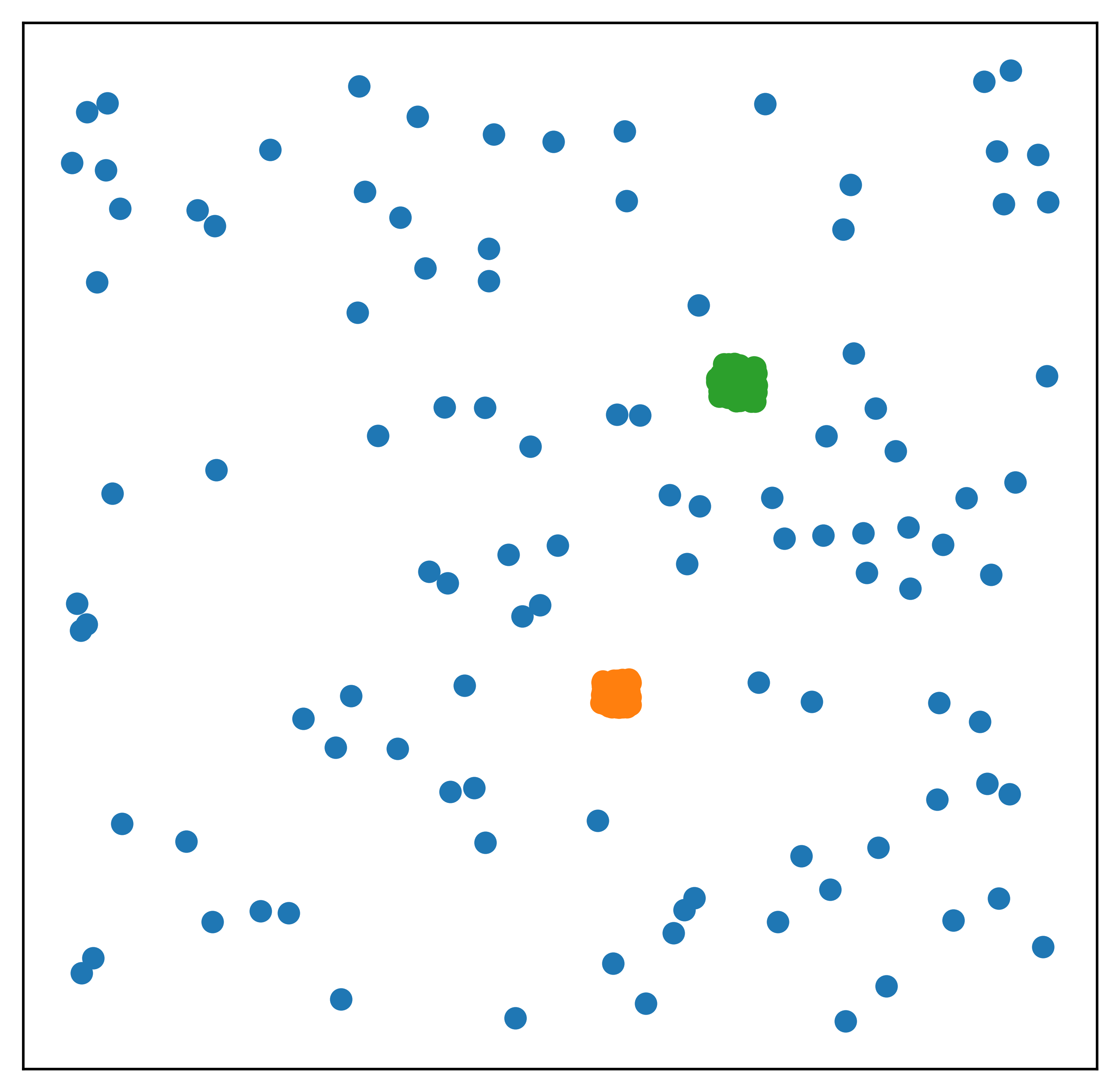}
		\caption{\texttt{sparse303}}
		\label{Fig:Fig-04-c}
	\end{subfigure}
	\begin{subfigure}{0.24\textwidth}
		\centering
		\includegraphics[width=\linewidth]{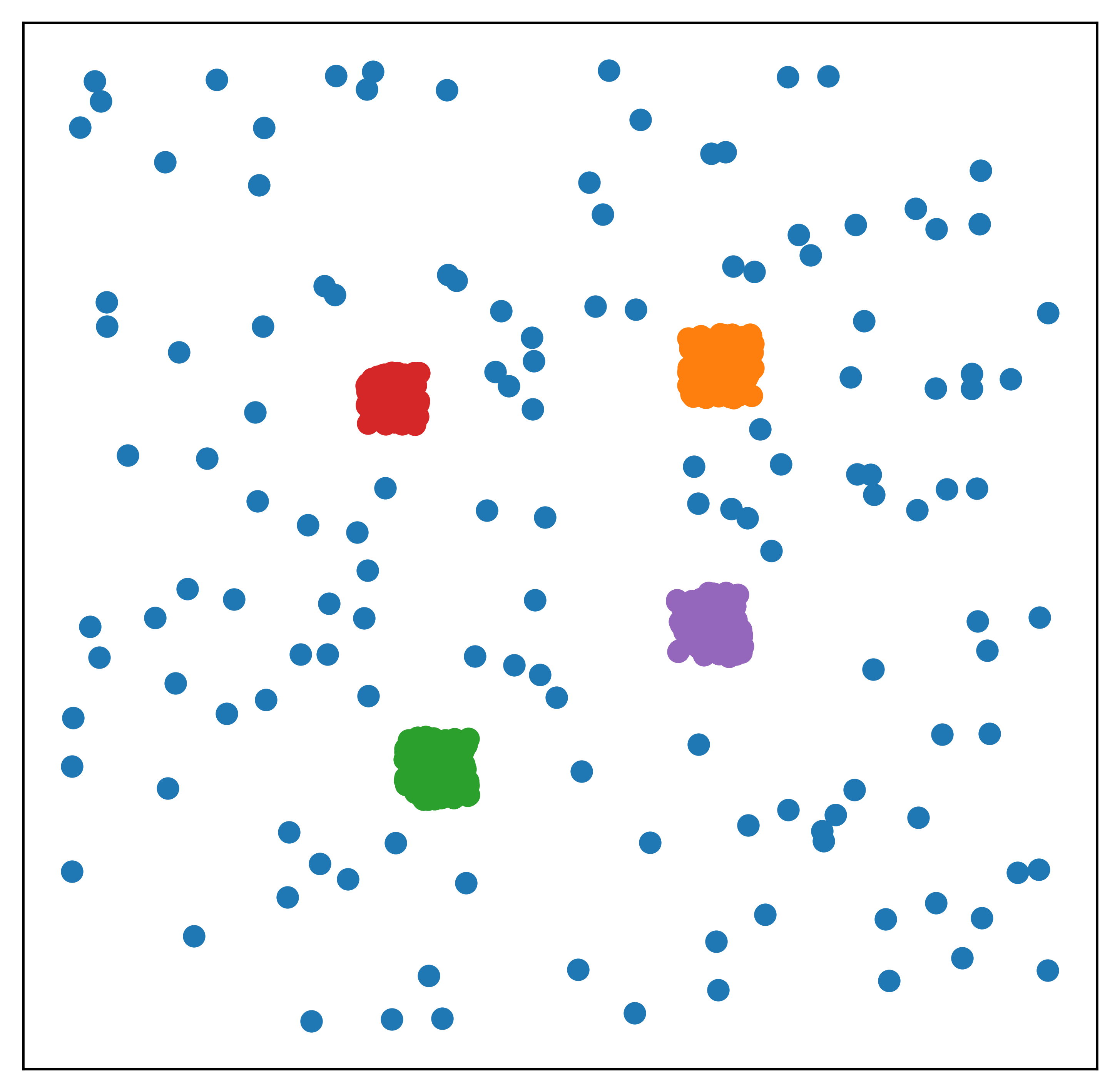}
		\caption{\texttt{sparse622}}
		\label{Fig:Fig-04-d}
	\end{subfigure}
	\caption{2-dimensional datasets used in the experiments. (Best viewed in color)}
	\label{Fig:2D-Datasets}
\end{figure}

Spectral analysis provides an elegant way to check the graph connectivity. The eigenvector $v_0$ associated with the smallest eigenvalue $\lambda_0$ of the graph Laplacian $L$, that eigenvector should be constant. This was stated in (Proposition 2) by von Luxburg \cite{Luxburg2007tutorial}. She wrote ``\textit{In a graph consisting of only one connected component we thus only have the constant one vector $\mathbbm{1}$ as eigenvector with eigenvalue 0}''. So, we used the standard deviation of points along the smallest eigenvector $v_0$ to see if the graph is connected or not. If the graph is connected (i.e. it contains a one connected component) the standard deviation will be small. We used the \texttt{ring238} dataset, which is a 2D dataset shown in Figure\ \ref{Fig:2D-Datasets}. In Figure\ \ref{Fig:EigenvectorStd}, there is a clear elbow point at $T=10$, which means the graph becomes connected from this point onwards.

The second problem related to the number of trees $T$ is setting $T$ to a large value, which affects the memory efficiency. Naturally if we start with a low $T$, some edges will be missing. These edges will be created as we increase $T$. At some point, the graph will have the same edges even if we increase $T$. Based on our empirical analysis we set $T=10$ in all experiments.

\subsection{GCN and LDS using $k$-nn and rpForest graphs}

In this experiment, we compared the performance of GCN and LDS using $k$-nn and rpForest graphs. We used four 2-dimensional datasets. These 2D datasets are shown in Figure\ \ref{Fig:2D-Datasets} with their class labels. We also used four datasets retrieved from scikit-learn library \cite{scikit-learn}. For train and test splits we used the same settings in LDS paper \cite{franceschi2019learning}. These settings are shown in Table\ \ref{Table:Datasets-Statistics}.

\begin{table}[]
	\caption{Summary statistics of the datasets.}
	\label{Table:Datasets-Statistics}
	\centering
	\includegraphics[width=0.60\textwidth,height=20cm,keepaspectratio]{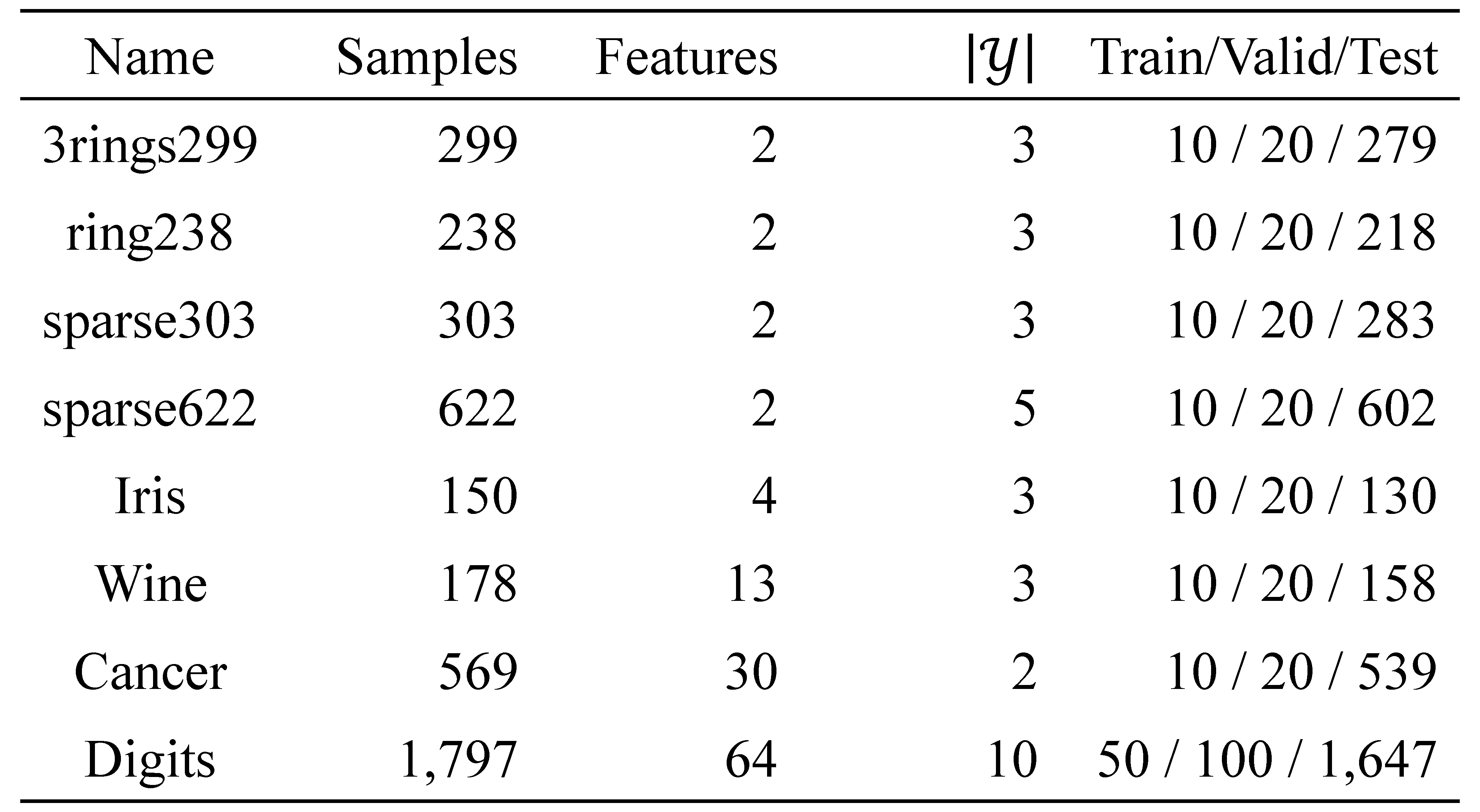}	
\end{table}

\begin{figure*}
	\centering
	\includegraphics[width=\textwidth]{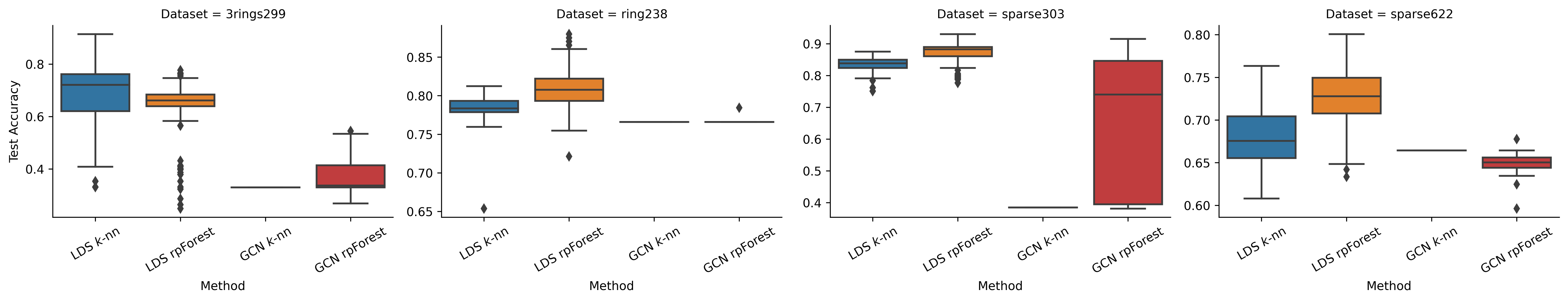}  
	\includegraphics[width=\textwidth]{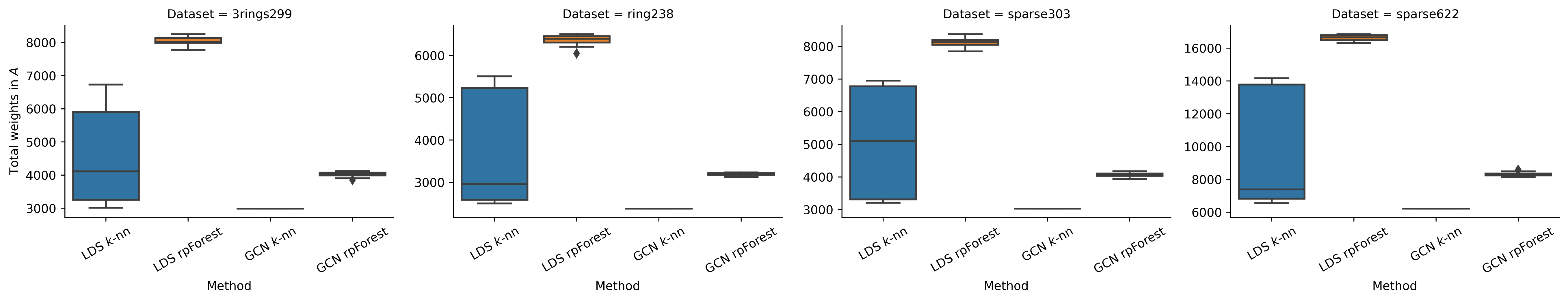}
	\caption{Running LDS and GCN using $k$-nn and rpForest graph on 2D datasets; (top) test accuracy; (bottom) total weights in the adjacency matrix $A$. (Best viewed in color)}
	\label{Fig:Results-2D}
\end{figure*}

Figure\ \ref{Fig:Results-2D} shows the results of running GCN and LDS on 2-dimensional datasets. In general, we can see LDS performed better than GCN, whether it is using $k$-nn graph or rpForest graph. This can be explained by how these two methods work. GCN takes the graph and runs it through the deep network, it cannot modify the graph by adding or removing edges. On the other hand, LDS uses the validation error to modify the graph by keeping the edges that minimize the validation error. Of course, LDS needs more time than GCN. 

LDS performed better when given a graph based on rpForest compared to $k$-nn graph. But this was not the case with \texttt{3rings299} dataset, when the linear line split by rpForest breaks the two rings in \texttt{3rings299} dataset. Also, we observed that rpForest graph has improved the performance of GCN. The total weights in the adjacency matrix $A$ gives us a hint about memory efficiency. Graphs in LDS have more weight than GCN, because LDS keeps modifying the graph by adding more edges. Another thing to highlight is rpForest graphs have more edges than $k$-nn graphs.

\begin{figure*}
	\centering
	\includegraphics[width=\textwidth]{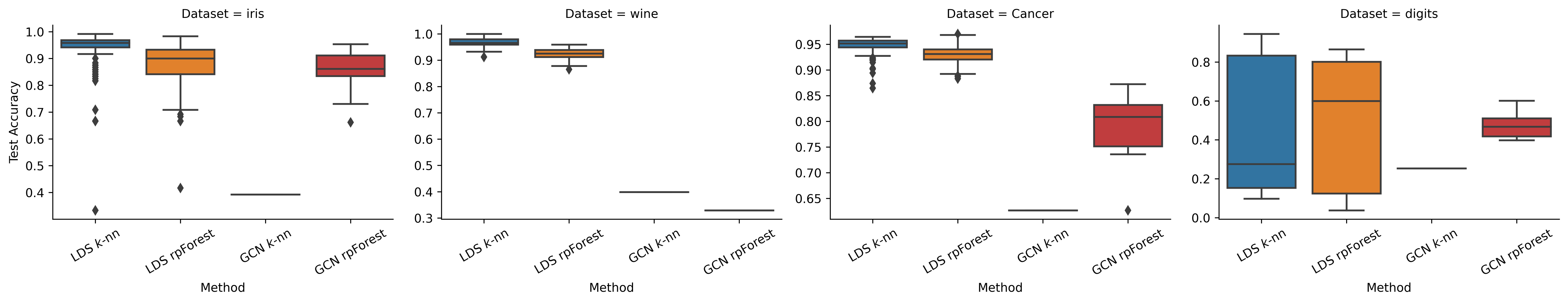}  
	\includegraphics[width=\textwidth]{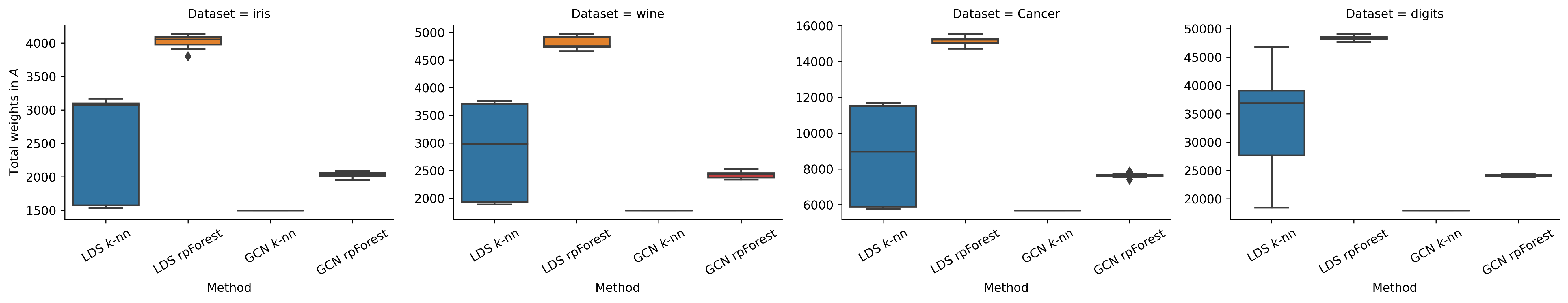}
	\caption{Running LDS and GCN using $k$-nn and rpForest graph on scikit-learn datasets; (top) test accuracy; (bottom) total weights in the adjacency matrix $A$. (Best viewed in color)}
	\label{Fig:Results-scikit}
\end{figure*}

The results of experiments on scikit-learn datasets are shown in Figure\ \ref{Fig:Results-scikit}. GCN test accuracy was very close to the one delivered by LDS in \texttt{iris} and \texttt{digits}, even though LDS has the ability to modify the graph. Another observation is that when a GCN is fed an rpForest graph it performs better compared to $k$-nn graph. For the total weights metric, we had the same observation across all datasets. LDS requires more storage especially when we feed it an rpForest graph, whereas GCN requires less storage.

\begin{figure}
	\centering
	\includegraphics[width=0.40\textwidth]{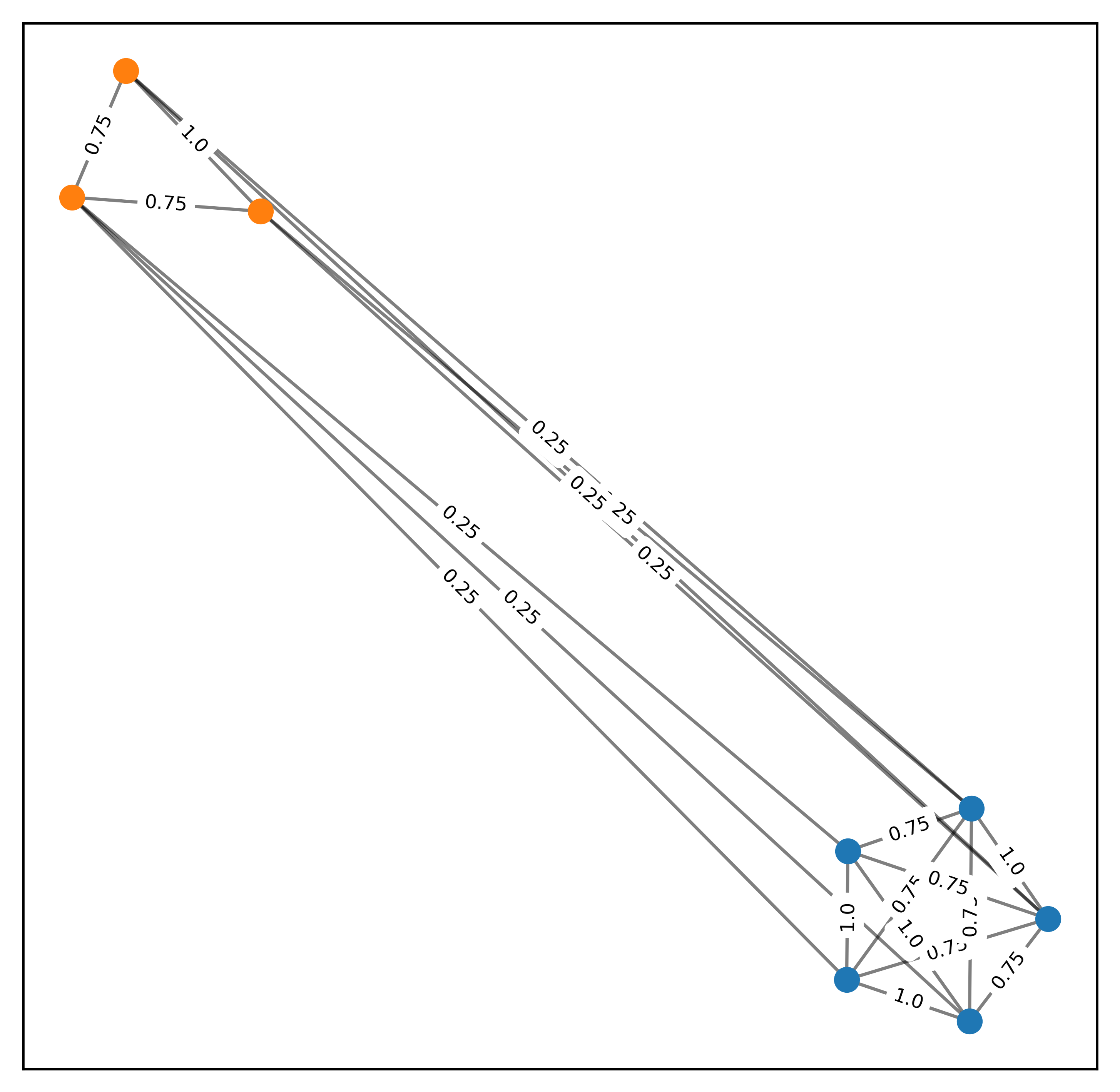}  
	\includegraphics[width=0.40\textwidth]{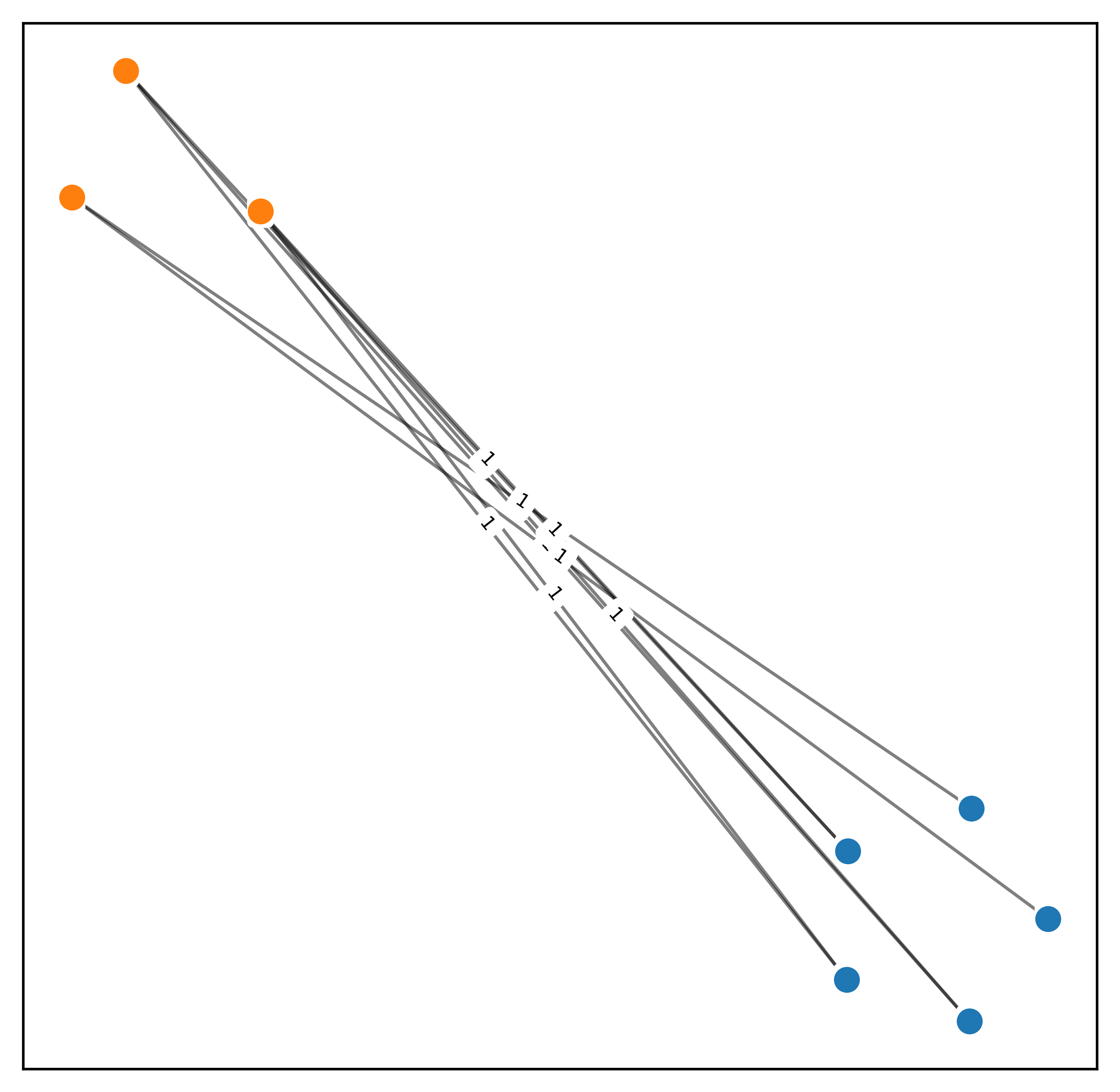}
	\caption{(left) an rpForest graph; (right) edges that did not appear in the rpForest graph. (Best viewed in color)}
	\label{Fig:graph-rpForest-Not}
\end{figure}

\subsection{LDS using rpForest graph with extra edges}

The rpForest graph connects only the points from the same leaf node. In this experiment, we investigate if we add some edges between points from different leaf nodes, would this improve the performance. Figure\ \ref{Fig:graph-rpForest-Not} shows an example of rpForest graph and the edges that did not appear in the rpForest graph. We want to examine if we take a percentage of these edges that did not appear in the rpForest graph, would that increase the connectivity and consequently improve the performance.

\begin{figure*}
	\centering
	\includegraphics[width=\textwidth]{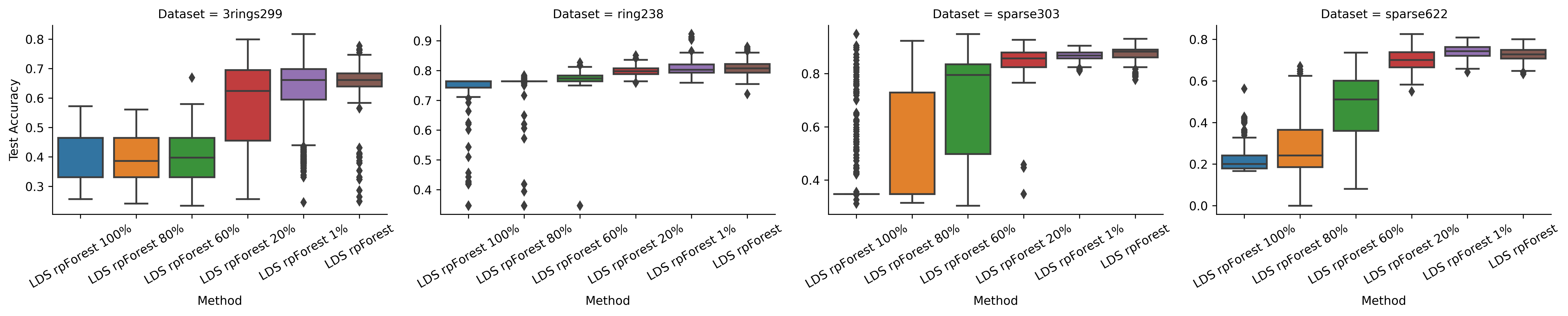}  
	\includegraphics[width=\textwidth]{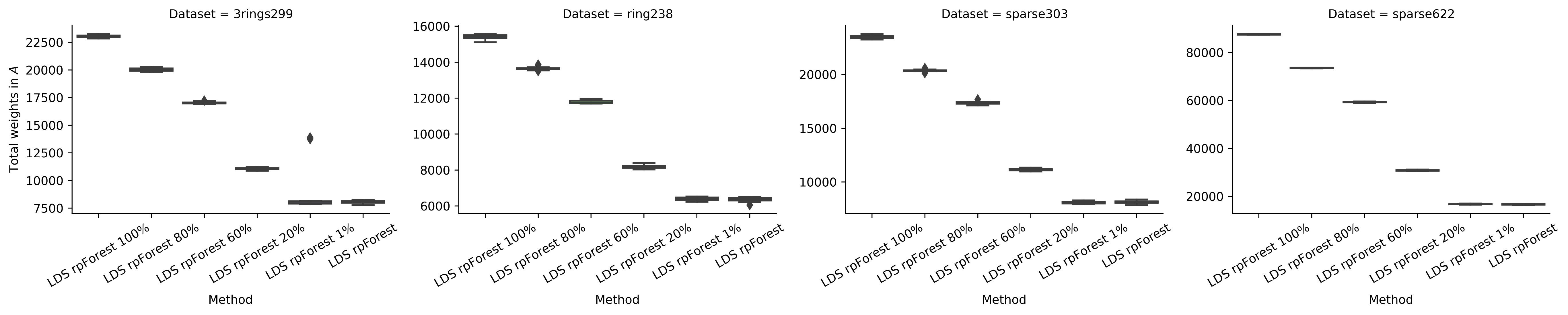}
	\caption{Running LDS using rpForest with extra edges; the percentage on the x-axis represents the percentage of extra edges; (top) test accuracy; (bottom) total weights in the adjacency matrix $A$. (Best viewed in color)}
	\label{Fig:Results-rpForest-percent}
\end{figure*}

By looking at Figure\ \ref{Fig:Results-rpForest-percent}, which shows LDS test accuracies using extra edges, we can see that these extra edges did not improve the performance. Even at the most extreme case when we included 100\% of these edges, the performance dropped by 50\% in some datasets. The memory footprint of these extra edges was very large. These findings emphasize on the ability of rpForest to find the most important edges for classification.

One advantage for our method is assigning weights proportionate to the edge's occurrence in the rpForest. This allows non-equal weights across the graph. A potential application for our method is the analysis of complex networks in the brain \cite{Stam2007Graph}. This is a research area in neuroscience that studies the complex connectivity on neuronal circuit dynamics. The functional connectivity between brain areas can be modeled as edges on a graph. These edges must have some varying weights, which is provided by our method.

\section{Conclusion}
\label{Conclusion}

Graphs are useful in modeling real-world relationships. That is why researchers were keen on extending deep learning to graphs. One of the successful applications of deep learning on graphs is graph convolutional networks (GCNs). The problem with GCN is that it needs the graph prepared beforehand. In most cases, the graph must be constructed from the dataset. A common choice to construct the graph is to use $k$-nearest neighbor graph. But $k$-nn assigns equal weights on all edges, which gives all edges the same importance during deep learning training.

We present a graph based on random projection forests (rpForest) with varying weights on edges. The weight on the edge was set proportional to how many trees it appears on. The number of trees is a hyperparameter in rpForest that needs careful tuning. We performed spectral analysis that helps us to set this parameter within the right range. The experiments revealed that initializing GCN using rpForest delivers better accuracy than $k$-nn initialization. We also showed that the edges provided by rpForest are the best for learning and adding extra edges did not improve the performance.

For future work, we can try a different weight assignment strategy other than average, a Euclidean distance for example. Another potential extension to our work could be investigating how different binary space-partitioning trees would affect the performance of the GCN. Also, it is important to examine how rpForest graph would perform in different methods of graph neural networks (GNNs).


\begin{thebibliography}{31}
	\providecommand{\natexlab}[1]{#1}
	\providecommand{\url}[1]{\texttt{#1}}
	\expandafter\ifx\csname urlstyle\endcsname\relax
	  \providecommand{\doi}[1]{doi: #1}\else
	  \providecommand{\doi}{doi: \begingroup \urlstyle{rm}\Url}\fi
	
	\bibitem[Belkin and Niyogi(2001)]{Belkin2001Laplacian}
	Mikhail Belkin and Partha Niyogi.
	\newblock Laplacian eigenmaps and spectral techniques for embedding and
	  clustering.
	\newblock \emph{Advances in neural information processing systems}, 14, 2001.
	
	\bibitem[Bentley(1975)]{Bentley1975Multidimensional}
	Jon~Louis Bentley.
	\newblock Multidimensional binary search trees used for associative searching.
	\newblock \emph{Commun. ACM}, 18\penalty0 (9):\penalty0 509–517, September
	  1975.
	\newblock \doi{https://doi.org/10.1145/361002.361007}.
	
	\bibitem[Bruna et~al.(2013)Bruna, Zaremba, Szlam, and LeCun]{Bruna2013Spectral}
	Joan Bruna, Wojciech Zaremba, Arthur Szlam, and Yann LeCun.
	\newblock Spectral networks and locally connected networks on graphs, 2013.
	
	\bibitem[Chen et~al.(2015)Chen, Liu, and Sun]{Chen2015Anomaly}
	Fan Chen, Zicheng Liu, and Ming-ting Sun.
	\newblock Anomaly detection by using random projection forest.
	\newblock In \emph{2015 IEEE International Conference on Image Processing
	  (ICIP)}, pages 1210--1214, 2015.
	\newblock \doi{https://doi.org/10.1109/ICIP.2015.7350992}.
	
	\bibitem[Cong et~al.(2022)Cong, Gupta, Neumann, de~Bayser, Steiner, and
	  Conch{\'u}ir]{cong2022prediction}
	Guojing Cong, Anshul Gupta, Rodrigo Neumann, Maira de~Bayser, Mathias Steiner,
	  and Breannd{\'a}n~{\'O} Conch{\'u}ir.
	\newblock Prediction of $\textrm{CO}_2$ adsorption in nano-pores with graph
	  neural networks, 2022.
	
	\bibitem[Dasgupta and Freund(2008)]{Dasgupta2008Random}
	Sanjoy Dasgupta and Yoav Freund.
	\newblock Random projection trees and low dimensional manifolds.
	\newblock In \emph{Proceedings of the Fortieth Annual ACM Symposium on Theory
	  of Computing}, STOC '08, page 537–546, New York, NY, USA, 2008. Association
	  for Computing Machinery.
	\newblock ISBN 9781605580470.
	\newblock \doi{https://doi.org/10.1145/1374376.1374452}.
	
	\bibitem[Dasgupta and Sinha(2015)]{Dasgupta2015Randomized}
	Sanjoy Dasgupta and Kaushik Sinha.
	\newblock Randomized partition trees for nearest neighbor search.
	\newblock \emph{Algorithmica}, 72\penalty0 (1):\penalty0 237–263, May 2015.
	\newblock ISSN 0178-4617.
	\newblock \doi{https://doi.org/10.1007/s00453-014-9885-5}.
	
	\bibitem[Defferrard et~al.(2016)Defferrard, Bresson, and
	  Vandergheynst]{Defferrard2016Convolutional}
	Michaël Defferrard, Xavier Bresson, and Pierre Vandergheynst.
	\newblock Convolutional neural networks on graphs with fast localized spectral
	  filtering, 2016.
	
	\bibitem[Franceschi et~al.(2019)Franceschi, Niepert, Pontil, and
	  He]{franceschi2019learning}
	Luca Franceschi, Mathias Niepert, Massimiliano Pontil, and Xiao He.
	\newblock Learning discrete structures for graph neural networks.
	\newblock In \emph{Proceedings of the 36th International Conference on Machine
	  Learning}, pages 1972--1982, 2019.
	
	\bibitem[Freund et~al.(2007)Freund, Dasgupta, Kabra, and
	  Verma]{Freund2008Learning}
	Yoav Freund, Sanjoy Dasgupta, Mayank Kabra, and Nakul Verma.
	\newblock Learning the structure of manifolds using random projections.
	\newblock \emph{Advances in Neural Information Processing Systems}, 20, 2007.
	
	\bibitem[Hammond et~al.(2011)Hammond, Vandergheynst, and
	  Gribonval]{Hammond2011Wavelets}
	David~K. Hammond, Pierre Vandergheynst, and Rémi Gribonval.
	\newblock Wavelets on graphs via spectral graph theory.
	\newblock \emph{Applied and Computational Harmonic Analysis}, 30\penalty0
	  (2):\penalty0 129--150, 2011.
	\newblock ISSN 1063-5203.
	\newblock \doi{https://doi.org/10.1016/j.acha.2010.04.005}.
	
	\bibitem[Henaff et~al.(2015)Henaff, Bruna, and LeCun]{Henaff2015Deep}
	Mikael Henaff, Joan Bruna, and Yann LeCun.
	\newblock Deep convolutional networks on graph-structured data, 2015.
	
	\bibitem[Keivani and Sinha(2021)]{Keivani2021Random}
	Omid Keivani and Kaushik Sinha.
	\newblock Random projection-based auxiliary information can improve tree-based
	  nearest neighbor search.
	\newblock \emph{Information Sciences}, 546:\penalty0 526--542, 2021.
	\newblock \doi{https://doi.org/10.1016/j.ins.2020.08.054}.
	
	\bibitem[Kipf and Welling(2017)]{kipf2017semi}
	Thomas~N. Kipf and Max Welling.
	\newblock Semi-supervised classification with graph convolutional networks,
	  2017.
	
	\bibitem[Li et~al.(2023)Li, Müller, Qian, Delgadillo, Abualshour, Thabet, and
	  Ghanem]{Li2023DeepGCNs}
	Guohao Li, Matthias Müller, Guocheng Qian, Itzel~C. Delgadillo, Abdulellah
	  Abualshour, Ali Thabet, and Bernard Ghanem.
	\newblock Deepgcns: Making gcns go as deep as cnns.
	\newblock \emph{IEEE Transactions on Pattern Analysis and Machine
	  Intelligence}, 45\penalty0 (6):\penalty0 6923--6939, 2023.
	\newblock \doi{https://doi.org/10.1109/TPAMI.2021.3074057}.
	
	\bibitem[Ng et~al.(2001)Ng, Jordan, and Weiss]{Ng2001Spectral}
	Andrew Ng, Michael Jordan, and Yair Weiss.
	\newblock On spectral clustering: Analysis and an algorithm.
	\newblock \emph{Advances in neural information processing systems}, 14, 2001.
	
	\bibitem[Pedregosa et~al.(2011)Pedregosa, Varoquaux, Gramfort, Michel, Thirion,
	  Grisel, Blondel, Prettenhofer, Weiss, Dubourg, Vanderplas, Passos,
	  Cournapeau, Brucher, Perrot, and Duchesnay]{scikit-learn}
	F.~Pedregosa, G.~Varoquaux, A.~Gramfort, V.~Michel, B.~Thirion, O.~Grisel,
	  M.~Blondel, P.~Prettenhofer, R.~Weiss, V.~Dubourg, J.~Vanderplas, A.~Passos,
	  D.~Cournapeau, M.~Brucher, M.~Perrot, and E.~Duchesnay.
	\newblock Scikit-learn: Machine learning in {P}ython.
	\newblock \emph{Journal of Machine Learning Research}, 12:\penalty0 2825--2830,
	  2011.
	
	\bibitem[Phan et~al.(2023)Phan, Nguyen, and Hwang]{Phan2023Aspect}
	Huyen~Trang Phan, Ngoc~Thanh Nguyen, and Dosam Hwang.
	\newblock Aspect-level sentiment analysis: A survey of graph convolutional
	  network methods.
	\newblock \emph{Information Fusion}, 91:\penalty0 149--172, 2023.
	\newblock \doi{https://doi.org/10.1016/j.inffus.2022.10.004}.
	
	\bibitem[Ram and Gray(2013)]{Ram2013Which}
	Parikshit Ram and Alexander Gray.
	\newblock Which space partitioning tree to use for search?
	\newblock \emph{Advances in Neural Information Processing Systems}, 26, 2013.
	
	\bibitem[Ren et~al.(2022)Ren, Lu, Xiao, Chang, Wang, Dong, and
	  Fang]{Ren2022Graph}
	Haotian Ren, Wei Lu, Yun Xiao, Xiaojun Chang, Xuanhong Wang, Zhiqiang Dong, and
	  Dingyi Fang.
	\newblock Graph convolutional networks in language and vision: A survey.
	\newblock \emph{Knowledge-Based Systems}, 251:\penalty0 109250, 2022.
	\newblock \doi{https://doi.org/10.1016/j.knosys.2022.109250}.
	
	\bibitem[Stam and Reijneveld(2007)]{Stam2007Graph}
	Cornelis~J Stam and Jaap~C Reijneveld.
	\newblock Graph theoretical analysis of complex networks in the brain.
	\newblock \emph{Nonlinear Biomedical Physics}, 2007.
	\newblock \doi{https://doi.org/10.1186/1753-4631-1-3}.
	
	\bibitem[Tavallali et~al.(2021)Tavallali, Tavallali, and
	  Singhal]{Tavallali2021Kmeans}
	Pooya Tavallali, Peyman Tavallali, and Mukesh Singhal.
	\newblock K-means tree: an optimal clustering tree for unsupervised learning.
	\newblock \emph{The Journal of Supercomputing}, 77, 2021.
	\newblock \doi{https://doi.org/10.1007/s11227-020-03436-2}.
	
	\bibitem[Tsitsulin et~al.(2023)Tsitsulin, Palowitch, Perozzi, and
	  M{\"u}ller]{Tsitsulin2023Graph}
	Anton Tsitsulin, John Palowitch, Bryan Perozzi, and Emmanuel M{\"u}ller.
	\newblock Graph clustering with graph neural networks.
	\newblock \emph{Journal of Machine Learning Research}, 24\penalty0
	  (127):\penalty0 1--21, 2023.
	\newblock URL \url{http://jmlr.org/papers/v24/20-998.html}.
	
	\bibitem[Veličković et~al.(2018)Veličković, Cucurull, Casanova, Romero,
	  Li{\'o}, and Bengio]{velickovic2018graph}
	Petar Veličković, Guillem Cucurull, Arantxa Casanova, Adriana Romero, Pietro
	  Li{\'o}, and Yoshua Bengio.
	\newblock Graph attention networks, 2018.
	
	\bibitem[von Luxburg(2007)]{Luxburg2007tutorial}
	Ulrike von Luxburg.
	\newblock A tutorial on spectral clustering.
	\newblock \emph{Statistics and Computing}, 17\penalty0 (4):\penalty0 395--416,
	  2007.
	\newblock ISSN 1573-1375.
	\newblock \doi{https://doi.org/10.1007/s11222-007-9033-z}.
	
	\bibitem[Wei et~al.(2023)Wei, Chen, Yin, Zhu, Zhou, and Liu]{Wei2023Adaptive}
	Lai Wei, Zhengwei Chen, Jun Yin, Changming Zhu, Rigui Zhou, and Jin Liu.
	\newblock Adaptive graph convolutional subspace clustering.
	\newblock In \emph{Proceedings of the IEEE/CVF Conference on Computer Vision
	  and Pattern Recognition (CVPR)}, pages 6262--6271, June 2023.
	
	\bibitem[Wu et~al.(2021)Wu, Pan, Chen, Long, Zhang, and
	  Yu]{Wu2021Comprehensive}
	Zonghan Wu, Shirui Pan, Fengwen Chen, Guodong Long, Chengqi Zhang, and
	  Philip~S. Yu.
	\newblock A comprehensive survey on graph neural networks.
	\newblock \emph{IEEE Transactions on Neural Networks and Learning Systems},
	  32\penalty0 (1):\penalty0 4--24, 2021.
	\newblock \doi{https://doi.org/10.1109/TNNLS.2020.2978386}.
	
	\bibitem[Yan et~al.(2018)Yan, Wang, Wang, Wang, and Li]{Yan2018Nearest}
	Donghui Yan, Yingjie Wang, Jin Wang, Honggang Wang, and Zhenpeng Li.
	\newblock K-nearest neighbor search by random projection forests.
	\newblock In \emph{2018 IEEE International Conference on Big Data (Big Data)},
	  pages 4775--4781, 2018.
	\newblock \doi{https://doi.org/10.1109/BigData.2018.8622307}.
	
	\bibitem[Yan et~al.(2021)Yan, Wang, Wang, Wang, and Li]{Yan2021Nearest}
	Donghui Yan, Yingjie Wang, Jin Wang, Honggang Wang, and Zhenpeng Li.
	\newblock K-nearest neighbor search by random projection forests.
	\newblock \emph{IEEE Transactions on Big Data}, 7\penalty0 (1):\penalty0
	  147--157, 2021.
	\newblock \doi{https://doi.org/10.1109/TBDATA.2019.2908178}.
	
	\bibitem[You et~al.(2020)You, Chen, Wang, and Shen]{You2020GCN}
	Yuning You, Tianlong Chen, Zhangyang Wang, and Yang Shen.
	\newblock L2-gcn: Layer-wise and learned efficient training of graph
	  convolutional networks.
	\newblock In \emph{Proceedings of the IEEE/CVF Conference on Computer Vision
	  and Pattern Recognition (CVPR)}, pages 2127--2135, June 2020.
	
	\bibitem[Zelnik-Manor and Perona(2004)]{Zelnik2004Self}
	Lihi Zelnik-Manor and Pietro Perona.
	\newblock Self-tuning spectral clustering.
	\newblock \emph{Advances in neural information processing systems}, 17, 2004.
	
	\end{thebibliography}

\end{document}